\documentclass{iopconfser}
\usepackage{comment}
\usepackage{svg}
\usepackage{graphicx}      
\usepackage{amsmath}
\usepackage{amssymb}
\usepackage{algorithm}
\usepackage{algorithmic}
\usepackage{caption}
\usepackage{subcaption}
\usepackage[title]{appendix}
\usepackage{multirow}
\usepackage{commath}
\usepackage{mathtools}
\usepackage{comment}
\usepackage{pdflscape}
\usepackage{soul}
\usepackage{adjustbox}
\usepackage{enumitem}
\usepackage{xcolor}
\usepackage{pifont}
\usepackage{todonotes}
\usepackage{lscape}
\usepackage{setspace}
\usepackage{hyperref}
\begin{document}

\title{COLREGs Compliant Collision Avoidance and Grounding Prevention for Autonomous Marine Navigation}

\author{Mayur S. Patil$^{1\dagger}$,
Nataraj Sudharsan$^{1\dagger}$,
Veneela Ammula$^{2}$,
Jude Tomdio$^{2}$,
Jin Wang$^{2}$,
Michael Kei$^{2}$,
Sivakumar Rathinam$^{1}$,
Prabhakar R. Pagilla$^{1}$
}

\affil{$^1$Department of Mechanical Engineering, Texas A\&M University, College Station, USA}


\affil{$^2$American Bureau of Shipping, Spring, TX, USA}

\affil{$^\dagger$Authors contributed equally}

\email{srathinam@tamu.edu}

\begin{abstract}

Maritime Autonomous Surface Ships (MASS) are increasingly regarded as a promising solution to address crew shortages, improve navigational safety, and improve operational efficiency in the maritime industry. Nevertheless, the reliable deployment of MASS in real-world environments remains a significant challenge, particularly in congested waters where the majority of maritime accidents occur. This emphasizes the need for safe and regulation-aware motion planning strategies for MASS that are capable of operating under dynamic maritime conditions. This paper presents a unified motion planning method for MASS that achieves real time collision avoidance, compliance with International Regulations for Preventing Collisions at Sea (COLREGs), and grounding prevention. The proposed work introduces a convex optimization method that integrates velocity obstacle-based (VO) collision constraints, COLREGs-based directional constraints, and bathymetry-based grounding constraints to generate computationally efficient, rule-compliant optimal velocity selection. To enhance robustness, the classical VO method is extended to consider uncertainty in the position and velocity estimates of the target vessel. Unnavigable shallow water regions obtained from bathymetric data, which are inherently nonconvex, are approximated via convex geometries using a integer linear programming (ILP), allowing grounding constraints to be incorporated into the motion planning. The resulting optimization generates optimal and dynamically feasible input velocities that meet collision avoidance, regulatory compliance, kinodynamic limits, and grounding prevention requirements. Simulation results involving multi-vessel encounters demonstrate the effectiveness of the proposed method in producing safe and regulation-compliant maneuvers, highlighting the suitability of the proposed approach for real time autonomous maritime navigation.

\end{abstract}

\section{Introduction}

The safe operation of MASS fundamentally depends on reliable navigation systems capable of operating effectively in complex maritime environments, including dense traffic scenarios, uncertain vessel dynamics, restricted waterways, and adverse weather conditions. Accident statistics show that collisions, groundings, loss of control, and contact events account for a large proportion of reported incidents in these dynamic situations~\cite{eliopoulou2023statistical}. Such operational risks emphasize the need for advanced motion planning methods that can proactively prevent collision and grounding while accounting for environmental uncertainty and navigational constraints. Safe autonomous navigation requires the generation of continuous and feasible waypoints that avoid collisions, comply with the COLREGs~\cite{imo1972colreg}, and respect environmental limitations such as shallow water regions and restricted channels. Compared to ground and aerial vehicles, marine vessels exhibit large inertia, slow response dynamics, and limited maneuverability, resulting in long planning horizons and reduced capability for rapid corrective actions. Consequently, motion planning for autonomous ships must anticipate potential hazards in advance and integrate collision and grounding avoidance with operational and environmental context to ensure safe and reliable navigation.


Numerous approaches have been proposed in the literature to address this problem. Optimization based strategies, particularly those built upon Model Predictive Control (MPC), explicitly account for vessel dynamics and operational constraints within a receding-horizon framework. While such methods provide systematic handling of constraints and predictive capability, their computational complexity can increase significantly with higher traffic density and longer prediction horizons~\cite{johansen2016ship, kufoalor2018colregs, eriksen2019model}. Furthermore, data-driven techniques, including deep reinforcement learning, have recently gained attention due to their ability to learn complex collision avoidance behaviors. Nevertheless, their reliance on large-scale training data and the limited transparency of learned policies present challenges for deployment in safety-critical maritime operations~\cite{penttinen2020colreg, sawada2021automatic, guan2023generalized}. Similarly, sampling-based planners such as A* and Rapidly-exploring Random Trees (RRT) enable trajectory generation in complex environments. However, these methods often require additional rule-based modules to ensure regulatory compliance, and their adaptability is limited under highly dynamic conditions~\cite{campbell2012rule}. These limitations have led to renewed interest in reactive approaches, notably Artificial Potential Fields (APF)~\cite{khatib1986real} and VO methods, which prioritize computational efficiency and real time feasibility while maintaining effective collision avoidance performance~\cite{fiorini1998motion}.

Although substantial research has focused on collision avoidance, grounding prevention remains relatively underexplored. In marine environment, bathymetric data provide areas of insufficient water depth where vessels are exposed to grounding hazards due to reduced under-keel clearance and shallow water effects such as squat. Effective motion planning must therefore extend beyond vessel-to-vessel interaction and explicitly account for grounding related hazards within the navigational environment. However, incorporating bathymetric constraints into real time planning presents significant challenges. Shallow water regions are typically characterized by irregular, nonconvex geometries, thus their direct integration in optimization-based planners is computationally demanding. This issue is particularly critical for approaches that depend on convex formulations to ensure computational efficiency, and real time feasibility. Therefore, developing a feasible real time methodology that accounts for both grounding prevention and dynamic obstacle collision avoidance while complying with COLREGs is essential for robust maritime motion planning.


Motivated by these challenges, this work proposes a unified and regulation-aware motion planning framework that simultaneously addresses collision avoidance, automatic COLREGs compliance, and grounding prevention under state uncertainty. The core contribution lies in formulating the motion planning problem as a computationally efficient convex optimization program in velocity space by efficiently integrating collision avoidance constraints and COLREGs-based directional rules as linear half-planes. To enable grounding prevention, non-convex shallow water regions derived from bathymetric data are convexified through geometric approximations using an ILP set-coverage formulation, allowing direct incorporation of depth-related safety constraints into the same optimization formulation. Within this structure, a robust reformulation of the classical VO method is introduced to explicitly account for bounded uncertainties in neighboring vessel position and velocity estimates, enhancing safety under realistic sensing conditions. The proposed unified formulation preserves computational efficiency while ensuring dynamically feasible and regulation-compliant velocity selection. The effectiveness of the proposed method is validated through simulations based on IMAZU encounter scenarios~\cite{patil2025virtual}. The results demonstrate that the approach computes dynamically feasible velocities that optimally balance goal-directed motion and safety requirements while ensuring collision avoidance, compliance with COLREGs, and grounding protection. These results highlight the robustness and practical viability of the method for autonomous real time navigation in congested and shallow water environments, where reliable and regulation-consistent navigation is critical. For a broader and more rigorous evaluation, the proposed method will be integrated into an advanced simulation framework currently under development through a collaborative initiative between Texas A\&M University and the American Bureau of Shipping (ABS)~\cite{patil2025virtual, patil2026robust}. Such integration would enable evaluation across diverse environmental conditions and operational complexities, further strengthening the method’s applicability to real-world deployment.

The remainder of the paper is organized as follows. Section 2 provides the essential background and preliminaries needed for the development of the paper. The methodology of the proposed formulation and its development are described in Section 3. A representative sample of simulation results with discussions that highlight the key features of the proposed method is given in Section 4. Conclusions and future work are given in Section 5.

\section{Background}

This section provides the fundamental background and preliminaries necessary for the formulation and analysis of the proposed collision avoidance and grounding prevention methodology.

\subsection{Classical Velocity Obstacle Algorithm}\label{sec:vo_algo}

The classical VO approach, introduced in~\cite{fiorini1998motion, fiorini1993motion}, provides a geometric approach to real time collision avoidance by operating directly in velocity space over position space. The VO identifies a subset of the velocity space that could lead to future collisions and selects admissible velocities outside these sets.

Consider two circular agents $A$ and $B$ with positions $p_A, p_B \in \mathbb{R}^2$, velocities $v_A, v_B$, and radii $r_A, r_B$. The relative position and velocity are defined as
\begin{align}
p_{AB} = p_B - p_A, \quad 
v_{AB} = v_A - v_B.
\end{align}

To construct the collision cone, agent $B$ is mapped into the configuration space of agent $A$ by reducing $A$ to a point $\hat{A}$ and expanding the geometry of $B$ by the radius of $A$, yielding the modified set $\hat{B}$. Based on this geometric construction, the collision cone $CC_{AB}$ is defined as:

\begin{equation}
CC_{AB} = \{\, v_{AB} \mid \lambda_{AB} \cap \hat{B} \neq \emptyset \,\}.
\label{eq:cc}
\end{equation}
where, $\lambda_{AB} = \{p_A + t v_{AB} \mid t\geq 0\}$ is a line of $v_{AB}$. Then the corresponding VO for $A$ induced by $B$, represented in terms of absolute velocity of $A$ is 
\begin{equation}
VO_{A|B} = \{\, v_A \mid v_{A} \in CC_{AB} \oplus v_B \,\}.
\label{eq:vo}
\end{equation}
where, $\oplus$ represents the Minkowski sum. All velocities of the agent $A$ inside $VO_{A|B}$ lead to collision under constant relative velocity assumptions, whereas velocities outside are guaranteed to be collision-free. This is generalized for multiple dynamic obstacles $\{B_i\}_{i=1}^n$, as follows:
\begin{equation}
\mathcal{V}_A^{\text{safe}} =
\mathbb{R}^2 \setminus \bigcup_{i=1}^n VO_{A|B_i}.
\end{equation}

The above formulation assumes instantaneous velocity changes. To incorporate dynamic constraints,~\cite{fiorini1998motion} introduced the concept of reachable velocities in VO. Let the feasible acceleration set be
\begin{equation}
FA(t) = \{\ddot{x} \mid \ddot{x} = f(x,\dot{x},u),\, u \in \mathcal{U}\},
\end{equation}
where $f(\cdot)$ represents the system dynamics and $\mathcal{U}$ the admissible control inputs. The reachable velocity set over a finite time interval $\Delta t$ is
\begin{equation}
RV(t+\Delta t) =
\{\, v_A(t) + \ddot{x}\Delta t \mid \ddot{x} \in FA(t) \,\}.
\end{equation}
The reachable avoidance velocity (RAV) set is then defined as
\begin{equation}
RAV(t+\Delta t) =
RV(t+\Delta t) \setminus VO(t),
\end{equation}
which represents set of instantaneous dynamically feasible and collision-free velocities.


\subsection{Collision Triggers: TCPA and DCPA}

In maritime collision risk assessment, the potential for a hazardous encounter between two vessels is commonly evaluated using two predictive measures: the \emph{Time to Closest Point of Approach} (TCPA) and the \emph{Distance at Closest Point of Approach} (DCPA). These measures describe the anticipated relative motion between the ego vessel and a target vessel, assuming steady-state kinematic conditions with constant velocities. By estimating both the temporal and spatial proximity at the closest approach, TCPA and DCPA provide a quantitative basis for evaluating collision risk. For a more detailed discussion including the formal mathematical derivation of these concepts, the readers are referred to \cite{kuwata2011safe, gai2024fuzzy}. 


\section{Methodology}

This section describes the proposed unified optimization-based motion planning approach for collision avoidance and grounding prevention in maritime navigation. The proposed approach computes a collision-free and COLREGs-compliant velocity using the Velocity Obstacle (VO) framework by solving a convex quadratic program (QP) in the velocity space that incorporates geometric collision-avoidance, regulatory requirements, and vessel maneuvering limits. In a deterministic formulation, the positions and velocities of the surrounding vessels are assumed to be accurately known, and environmental constraints are represented using hyperplanes. Under these assumptions, the admissible velocity region is convex, allowing the safe velocity to be efficiently computed through a convex QP.

To address practical operating conditions, the approach is extended to account for both sensing uncertainty and complex environmental geometry. Uncertainty in the estimated position and velocity of surrounding vessels is incorporated to construct a robust VO cone. In addition, shallow water regions are included as environmental constraints; when such regions are highly nonconvex, they are approximated using convex primitives to enable seamless integration into the optimization framework as static obstacles.

\begin{figure}[h]
\centering
\includegraphics[width=0.45\textwidth]{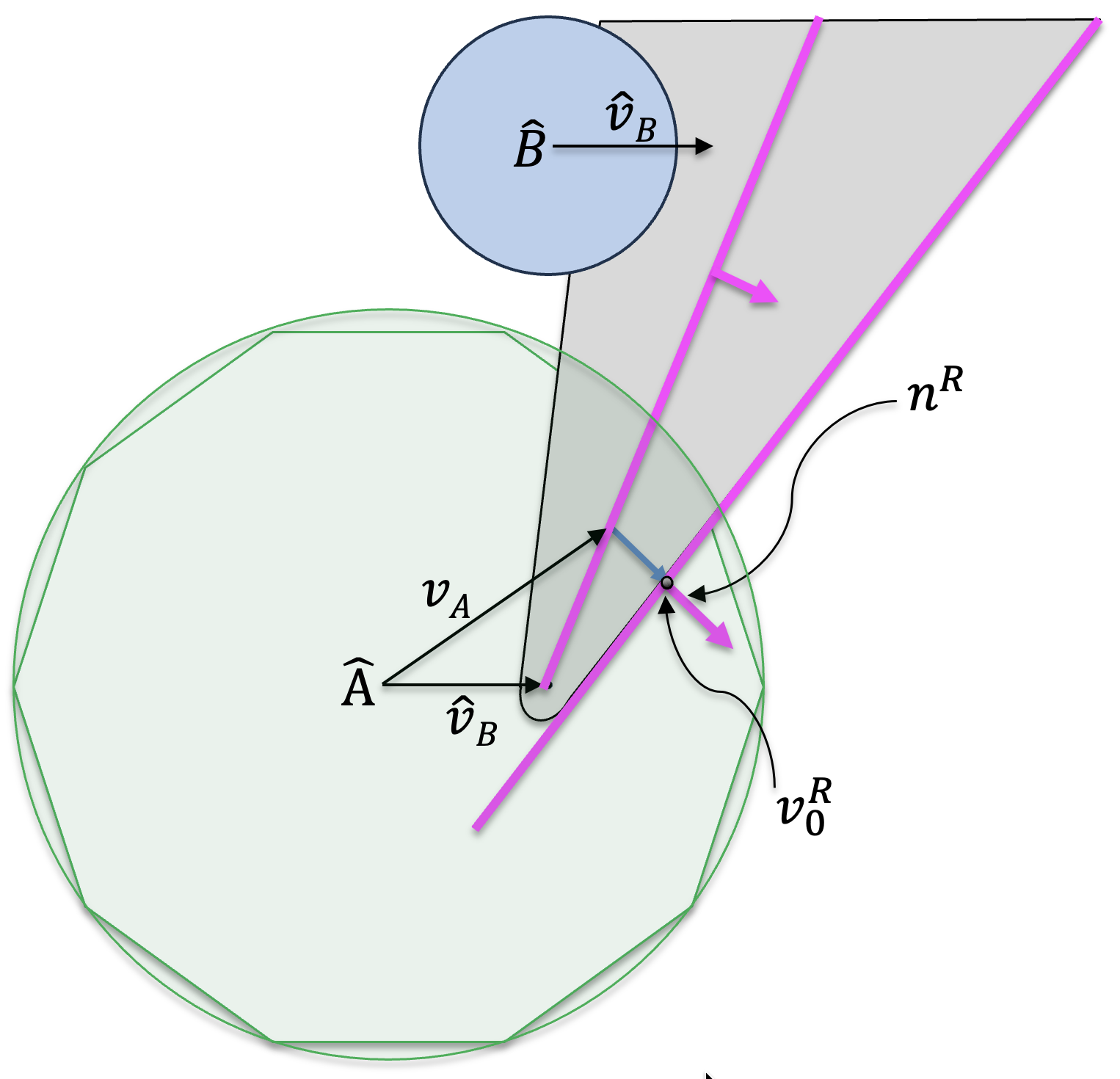}
\caption{Geometric interpretation of QP-based safe velocity selection in velocity space.}
\label{fig:vo_qp_figure}
\end{figure}

The deterministic QP formulation used to compute the safe velocity is described below. Within the VO framework (Fig.~\ref{fig:vo_qp_figure}), each surrounding vessel defines a region in the velocity space that would result in a future collision if selected by the ego vessel. Velocity collision cones are constructed only for vessels whose predicted TCPA and DCPA fall below predefined safety thresholds. The safe velocity is then obtained by solving a constrained optimization problem whose feasible set is defined by collision-avoidance, COLREGs, and physical maneuvering constraints, formulated as linear inequality constraints.

\paragraph{Collision-avoidance and COLREGs constraints:}
For each interacting vessel $i$, the corresponding velocity obstacle $VO_i$ is constructed. The boundary $\partial VO_i$ yields two candidate supporting half-planes associated with the left and right tangents. Let $(n_i^{L},v_{0,i}^{L})$ and $(n_i^{R},v_{0,i}^{R})$, with $v_{0,i}^{L},v_{0,i}^{R}\in\partial VO_i$, denote the outward unit normals and boundary points defining these two supporting half-planes.

The collision-avoidance constraint contributed by vessel $i$ is expressed as the selected supporting half-space
\begin{equation}
(n_i^{\sigma_i})^\top (v - v_{0,i}^{\sigma_i}) \ge 0,
\end{equation}
which is written in standard linear form as
\begin{equation}
-(n_i^{\sigma_i})^\top v \le -(n_i^{\sigma_i})^\top v_{0,i}^{\sigma_i}.
\end{equation}

The side index $\sigma_i \in \{L,R\}$ is determined based on the encounter geometry and COLREGs requirements:
\begin{equation}
\sigma_i =
\begin{cases}
R, & \text{if a starboard-side maneuver is required},\\
L, & \text{if a port-side maneuver is required},\\
\displaystyle \arg\max_{\sigma\in\{L,R\}} \; (n_i^{\sigma})^\top\!\left(v_{\mathrm{curr}}-v_{0,i}^{\sigma}\right), & \text{otherwise},
\end{cases}
\end{equation}
where $v_{\mathrm{curr}}$ denotes the current velocity.

\paragraph{Speed constraints:}
The vessel speed is limited by $|v| \le v_{\max}$. To retain a linear-constraint QP formulation, this circular bound is approximated by an inscribed polygon:
\begin{equation}
\tilde n_k^\top v \le \tilde c_k,
\qquad k = 1,\ldots,N_{\mathrm{poly}},
\end{equation}
where $N_{\mathrm{poly}}$ determines the approximation accuracy.

Stacking the VO-based constraints, COLREGs directional constraints, and polygonal speed limits yields a convex feasible set
\begin{equation}
\mathcal{V}_{\mathrm{feasible}} = \{ v \in \mathbb{R}^2 \mid Av \le b \}.
\end{equation}

\paragraph{Objective function:}
Among all feasible velocities in $\mathcal{V}_{\mathrm{feasible}}$, the following quadratic optimization problem yields the velocity that is closest to a desired reference velocity $v_{\mathrm{ref}}$:
\begin{equation}
\min_{v} \; \|v - v_{\mathrm{ref}}\|_{2}.
\end{equation}
The reference velocity represents the nominal motion of the ego vessel in the absence of conflicts and can correspond to a predefined velocity profile as well. The resulting QP with linear inequality constraints is solved using MATLAB's \texttt{quadprog}, yielding the safe velocity $v^\ast$. If the environment becomes overly constrained such that no feasible solution exists, the optimal action is to reduce speed and bring the vessel to a stop, by setting $v^\ast=0$.

\paragraph{Acceleration-limited update:}
Since the vessel cannot instantly realize optimal safe velocity $v^\ast$, the commanded velocity is updated subject to a maximum allowable change per time step. Let $v_{\mathrm{curr}}$ denote the current velocity and $d_{\max}$ the increase in maximum velocity. The applied velocity is given by 
\begin{equation}
v_{\mathrm{next}} =
\begin{cases}
v_{\mathrm{curr}} + 
d_{\max}\dfrac{v^\ast - v_{\mathrm{curr}}}{\|v^\ast - v_{\mathrm{curr}}\|}, 
& \|v^\ast - v_{\mathrm{curr}}\| > d_{\max}, \\[6pt]
v^\ast, & \|v^\ast - v_{\mathrm{curr}}\| \le d_{\max}.
\end{cases}
\end{equation}

\subsection{Collision Cone under Uncertainty}
\label{sec:uncertainty}

The VO constructed using Eq. \eqref{eq:cc} and \eqref{eq:vo} assumes perfect knowledge of the obstacle’s state, i.e., the position and velocity of agent $B$ are known precisely. However, in real-world scenarios, these states are obtained through radar and other onboard sensing modalities, which are inherently affected by estimation errors, environmental disturbances, and measurement noise. Consequently, the assumption of exact state information is unrealistic. To ensure reliable collision avoidance under such conditions, it is essential to formulate a robust VO that explicitly accounts for state uncertainties. Although \cite{kuwata2011safe} extends the VO method to incorporate the velocity uncertainty, the influence of position uncertainty has not been addressed in the literature. The approach in this paper fills this gap by extending the VO formulation to explicitly account for both position and velocity uncertainties.

\subsubsection*{Position Uncertainty:}

Let $\hat{p}_B \in \mathbb{R}^2$ denote the estimated position of agent $B$, and let $\delta_p \in \mathcal{U}_p$ represent a bounded uncertainty in this estimate. The true position of agent $B$ is therefore expressed as
\[
p_B = \hat{p}_B + \delta_p.
\]
By incorporating this uncertainty directly into the geometric construction of the collision cone (Eq.~\eqref{eq:cc}), we obtain the positional uncertainty-aware collision cone:
\begin{equation}
CC_{AB}^{\mathcal{U}} = \left\{ v_{AB} \;\middle|\; \lambda_{AB} \cap \left( \hat{B} \oplus \mathcal{U}_p \right) \neq \emptyset \right\}.
\label{eq:ucc}
\end{equation}
Geometrically, this formulation corresponds to the representation of agent $B$ in the configuration space of agent $A$ after inflating its geometry by both the shape of $A$ and the bounded uncertainty set $\mathcal{U}_p$. The effect of this inflation is an enlargement of the collision region, specifically through increased internal angle of the collision cone. Consequently, a broader set of relative velocities is classified as potentially collision prone, reflecting a conservative characterization of collision-inducing relative velocities.

\subsubsection*{Velocity Uncertainty:}

We also account for the uncertainty in the velocity estimate of agent $B$ following the methodology proposed in \cite{kuwata2011safe}. Let $\hat{v}_B \in \mathbb{R}^2$ denote the estimated velocity and $\delta_v \in \mathcal{U}_v$ a bounded velocity uncertainty of agent $B$, such that
\[
v_B = \hat{v}_B + \delta_v.
\]
Substituting this representation into the VO definition (Eq.~\eqref{eq:vo}) leads to the robust VO expressed in terms of the absolute velocity of agent $A$:
\begin{equation}
VO_{A|B}^{\mathcal{U}} = \left\{ v_A \;\middle|\; v_A \in CC_{AB}^{\mathcal{U}} \oplus \hat{v_B} \oplus \mathcal{U}_v \right\}.
\label{eq:uvo}
\end{equation}

\begin{figure}[h!]
    \centering
    \includegraphics[width=0.8\textwidth]{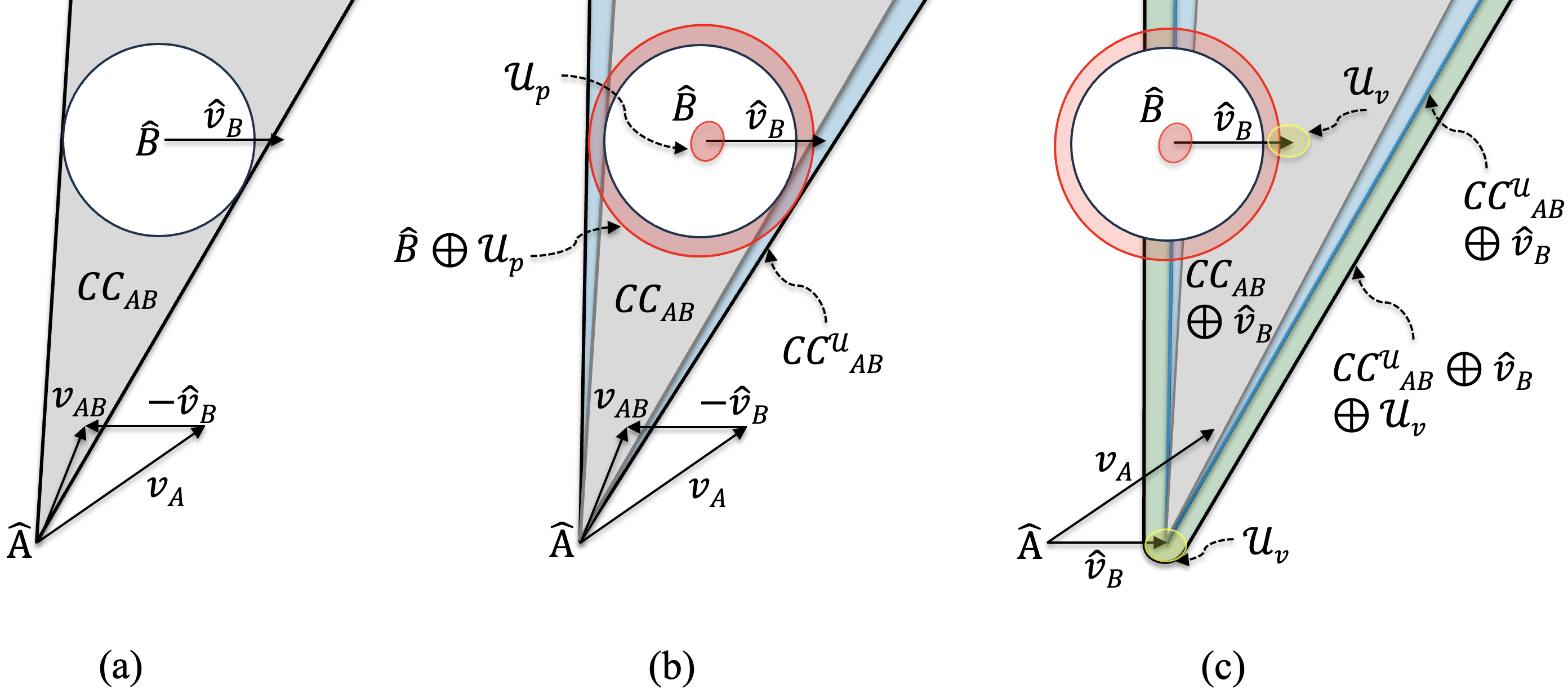}
    \caption{Geometric representation of collision cones. (a) Standard collision cone derived under perfect state information, (b) Expanded collision cone accounting for bounded position uncertainty of agent $B$, (c) Robust collision cone reflecting the combined effects of bounded position and velocity uncertainties.}
    \label{fig:cc_cone_construction}
\end{figure}

This construction can be interpreted as a Minkowski expansion of the VO by the bounded set $\mathcal{U}_v$, resulting in an enlarged VO region. The expansion directly reflects the worst-case position and velocity estimation error, thereby ensuring robustness against bounded disturbances. Figure~\ref{fig:cc_cone_construction} illustrates the geometric effect of incorporating both position and velocity uncertainties in the resulting VO cone.

\subsection{Convexification of shallow water Regions}\label{sec:convexification}

The proposed methodology approximates nonconvex shallow water regions using a set of convex geometric primitives, specifically circles. In VO-based collision avoidance methods, obstacles are commonly represented as circular discs. Therefore, modeling shallow water regions using circles enables direct integration into the existing VO formulation without requiring modifications to the underlying collision avoidance structure.

The primary objective is to construct an approximation that completely covers each shallow water region while minimizing both the number of circles employed and the inclusion of extraneous navigable space within the approximation. This requires optimizing both the placement and the radii of the circles such that their union approximates the target region with minimal over-approximation. This formulation corresponds to a variant of the classical set cover problem, which is known to be NP-hard, particularly when continuous circle placements are permitted. As a result, exact solutions are computationally intractable for large instances, necessitating the use of approximation methods, heuristics, or relaxation techniques. To this end, we formulate this problem as a discrete optimization problem by restricting the positions of the circle centers and their radii to finite candidate sets. This discretization allows the problem to be expressed as an integer linear program (ILP), which can be efficiently solved using standard solvers.



\emph{Problem Setup:}
Let $\mathcal{P} \subset \mathbb{R}^2$ denote a nonconvex planar polygonal region representing the shallow water area to be approximated. Consider a finite set of candidate circle centers defined as
\begin{equation}
\mathcal{C} = \{ c_i \in \mathbb{R}^2 \mid i = 1,\dots,M \},
\end{equation}
and a discrete set of allowable radii specified as
\begin{equation}
\mathcal{R} = \{ R_k \mid k = 1,\dots,K \}.
\end{equation}
To evaluate coverage, a uniform grid of points is sampled over a tight rectangular bounding box of $\mathcal{P}$, producing two point sets:
\begin{equation}
\begin{aligned}
\mathcal{X}_{\mathrm{in}} &= \{ p_j : p_j \in \mathrm{interior}(\mathcal{P}) \},\\
\mathcal{X}_{\mathrm{out}} &= \{ q_l : q_l \notin \mathcal{P} \}.
\end{aligned}
\end{equation}
Here, the interior points are mandatory coverage locations, while the exterior points quantify the undesired spillover region. For each point on the interior grid $p_j$, we precompute its coverage by each candidate circle:
\begin{equation}
A_{\mathrm{in}}(j,(i,k)) =
\begin{cases}
1, & \| p_j - c_i \| \le R_k, \\[3pt]
0, & \text{otherwise}.
\end{cases}
\end{equation}
Similarly, for each exterior point $q_l$,
\begin{equation}
A_{\mathrm{out}}(l,(i,k)) =
\begin{cases}
1, & \| q_l - c_i \| \le R_k, \\[3pt]
0, & \text{otherwise}.
\end{cases}
\end{equation}
These binary matrices encode the coverage relationship between candidate circles and spatial points, forming the foundation of the associated linear constraints and objective function. Furthermore, each candidate circle is assigned with a binary decision variable $z_{ik} \in \{0,1\}$ indicating whether the circle centered at $c_i$ with radius $R_k$ is selected in the final approximation. The discrete coverage problem is then formulated as follows:

\begin{equation}
\begin{aligned}
\min_{z} \quad &
\alpha \sum_{i=1}^{M} \sum_{k=1}^{K} z_{ik}
\;+\;
\beta \sum_{i=1}^{M} \sum_{k=1}^{K} \sum_{l=1}^{N_{\mathrm{out}}} A_{\mathrm{out}}(l,(i,k))\, z_{ik} \\[6pt]
\text{s.t.} \quad &
\sum_{i=1}^{M} \sum_{k=1}^{K}
A_{\mathrm{in}}(j,(i,k))\, z_{ik} \ge 1,
\qquad \forall j = 1,\dots,N_{\mathrm{in}}, \\[4pt]
&
z_{ik} \in \{0,1\},
\qquad \forall i = 1,\dots,M,\; \forall k = 1,\dots,K.
\end{aligned}
\end{equation}


The coefficients $\alpha$ and $\beta$ are tuning parameters for (i) the penalty associated with selecting additional circles and (ii) the penalty associated with over-approximating the shallow water region, respectively. Appropriate tuning of these parameters can ensure a balance between the number of circles and over-approximation error. Solving this optimization problem yields the optimal set of circles
\begin{equation}
\mathcal{C}^{\star}
= \{ (c_i, R_k) \mid z_{ik} = 1 \},
\end{equation}
which provides a multi-radius convex decomposition of the nonconvex shallow water region $\mathcal{P}$.

\begin{figure}[tbh!]
    \centering
    \includegraphics[width=0.45\textwidth]{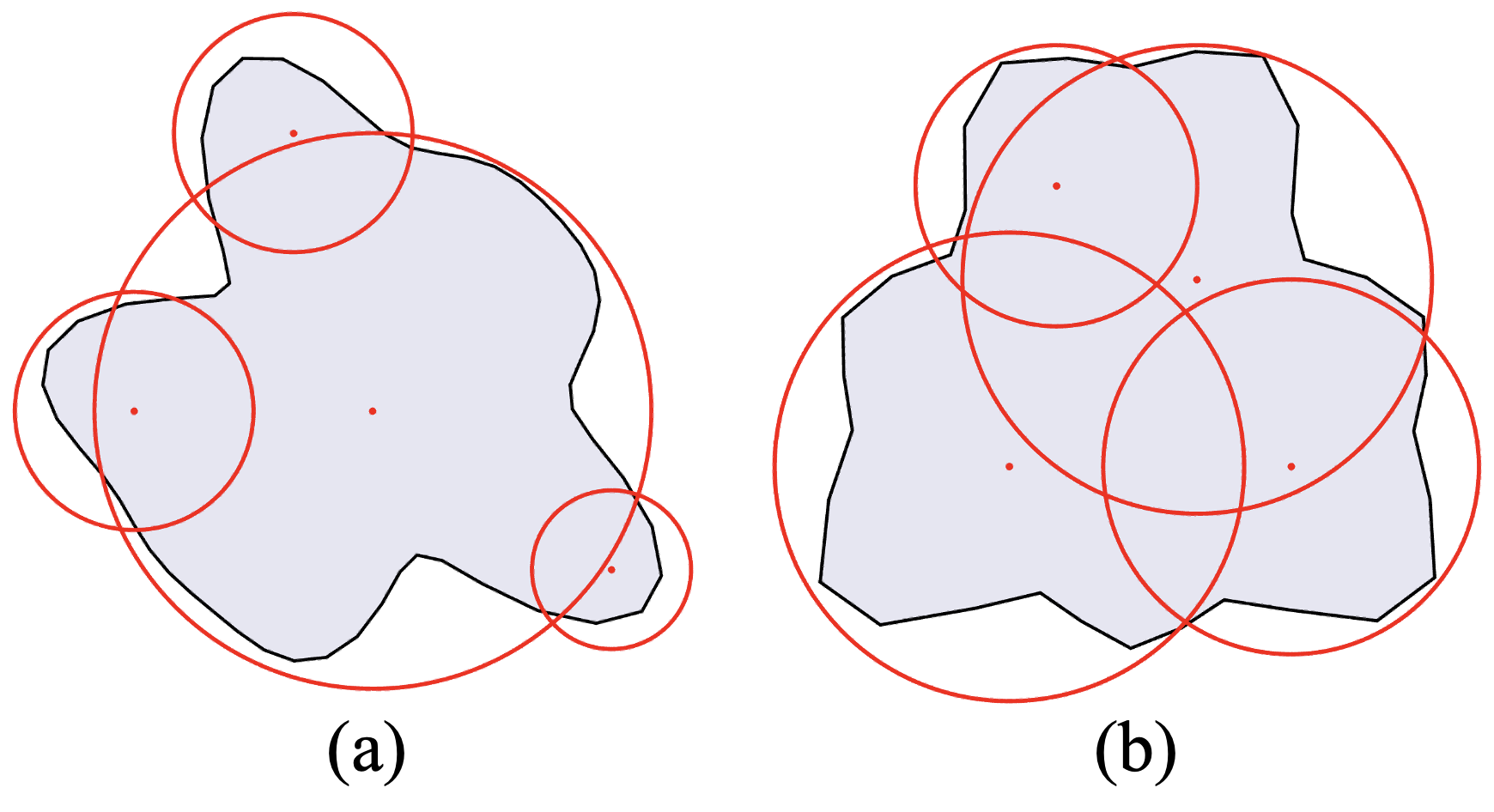}
    \caption{ILP results for convexification of nonconvex geometries}
    \label{fig:bathy_result}
\end{figure}

Figure~\ref{fig:bathy_result} illustrates the circle-cover approximations obtained for two representative nonconvex shallow water regions. 
In both cases, a uniform sampling grid with a resolution of $25\,\mathrm{m}$ was applied to the domain to obtain the discrete interior and exterior point sets $\mathcal{X}_{\mathrm{in}}$ and $\mathcal{X}_{\mathrm{out}}$, respectively. The candidate circle centers $\mathcal{C}$ were defined on a coarser grid $100\,\mathrm{m}$ across the same domain, yielding $M = 1681$ potential center locations. A discrete radius set $\mathcal{R} = \{100, 200, \dots, 1200\}\,\mathrm{m}$ (with $K = 12$ radii), and weighting coefficients $\alpha = 10$ and $\beta = 0.1$ were employed to enable flexible multi-scale coverage. 


In Fig.~\ref{fig:bathy_result} (a), the ILP solver returned a solution in 2.81\,\text{s}, requiring 135 iterations. The resulting approximation consists of four circles with radii 300, 300, 700, and 200\,\text{m}, respectively. The corresponding over approximation error, defined as the ratio of the difference of the circle cover area and the nonconvex shallow region area to the nonconvex shallow region area, was approximately 33\%.

In Fig.~\ref{fig:bathy_result} (b), the solver obtained an optimal solution in 2.26\,\text{s} with 87 iterations. Similar to case (a), four circles were selected, with radii 500, 400, 300, and 500\,\text{m}. The associated over approximation error in this case was approximately 30\% of the nonconvex shallow region area. This relatively high approximation errors are mainly due to the small size of the representative geometries and the coarse $100\,\mathrm{m}$ sampling resolution used for the candidate circle centers. In practical maritime settings, where shallow water regions are typically larger, a lower relative spillover error is expected with the same discrete set of radii. This trend is confirmed by refining the sampling resolution to $40\,\mathrm{m}$, which reduced the approximation error to 23\% and 25\% for cases (a) and (b), respectively. Further improvements can be achieved using denser sampling and a larger set of allowable radii.

Note that this convexification of nonconvex shallow water regions can be performed a priori to obtain an optimal set of circles, $\mathcal{C}^{\star}$ for known bathymetric regions or implemented in real time within a dynamically defined region of interest. This region can be determined using the reachable set of the vessel over a finite prediction horizon or through the use of TCPA/DCPA-based spatial limits to restrict the approximation domain.

\section{Simulation Results and Discussion}

The performance of the proposed approach is assessed under representative maritime scenarios adapted from the IMAZU benchmark test cases~\cite{patil2025virtual}. Specifically, the evaluation considers (i) a single-vessel head-on encounter and (ii) a multi-vessel scenario combining head-on and crossing encounters, both conducted in the presence of bathymetric shallow water regions. Since COLREGs do not explicitly define the angular threshold for head-on encounters, prior studies have adopted values such as $\pm 15^\circ$ or $\pm 5^\circ$. Here, a $\pm 5^\circ$ criterion is employed. Furthermore, strictly nonconvex shallow water regions were procedurally generated using a randomized Fourier perturbed radial model and placed within the region of interest to construct the test scenarios. The complete configuration details for the evaluated scenarios are summarized in Table~\ref{tab:scenario_settings}.



\begin{table}[ht!]
\centering
\caption{Initial conditions of target vessels for four encounter scenarios.}
\label{tab:scenario_settings}

\setlength{\tabcolsep}{2.5pt}
\renewcommand{\arraystretch}{1.5}
\footnotesize   

\begin{tabular}{|c|c|c|c|c|}
\hline
\textbf{Case} & \textbf{Vessels} & \textbf{Start (N,E) [m,m]} & \textbf{Heading [$^\circ$]} & \textbf{Speed [m/s]} \\
\hline
\multirow{1}{*}{(a)}
& V1 & (6000, 0) & $180$ & 10 \\
\hline
\multirow{2}{*}{(b)}
& V1 & (3500, 2500) & $270$ & 10 \\
\cline{2-5}
& V2 & (6000, 0) & $180$ & 10 \\
\hline

\end{tabular}
\end{table}

In all simulations, the ego vessel was initialized at $(0,0)$ with a nominal cruising speed of $10\,\mathrm{m/s}$. The control framework operated at $10\,\mathrm{Hz}$ ($\Delta t = 0.1\,\mathrm{s}$) over a fixed simulation horizon of $1000\,\mathrm{s}$, resulting in $10{,}000$ optimization iterations. The designated goal position was $(7000,0)$.


\begin{figure}[t!]
    \centering
    \subfloat[Head-on scenario]{%
        \includegraphics[width=0.48\linewidth]{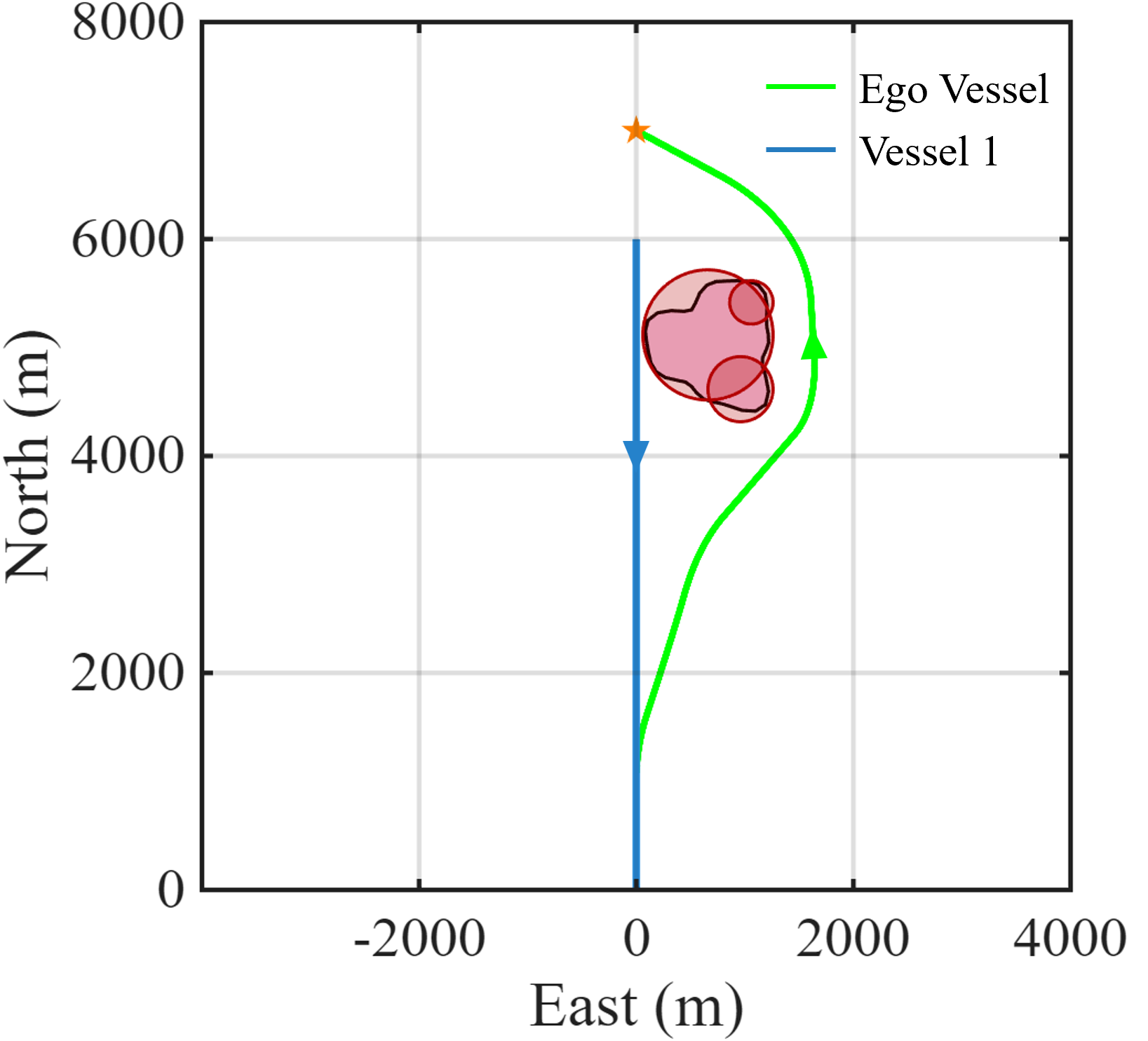}%
        \label{fig:result1}
    }
    \hfill
    \subfloat[Head-on and crossing scenario]{%
        \includegraphics[width=0.48\linewidth]{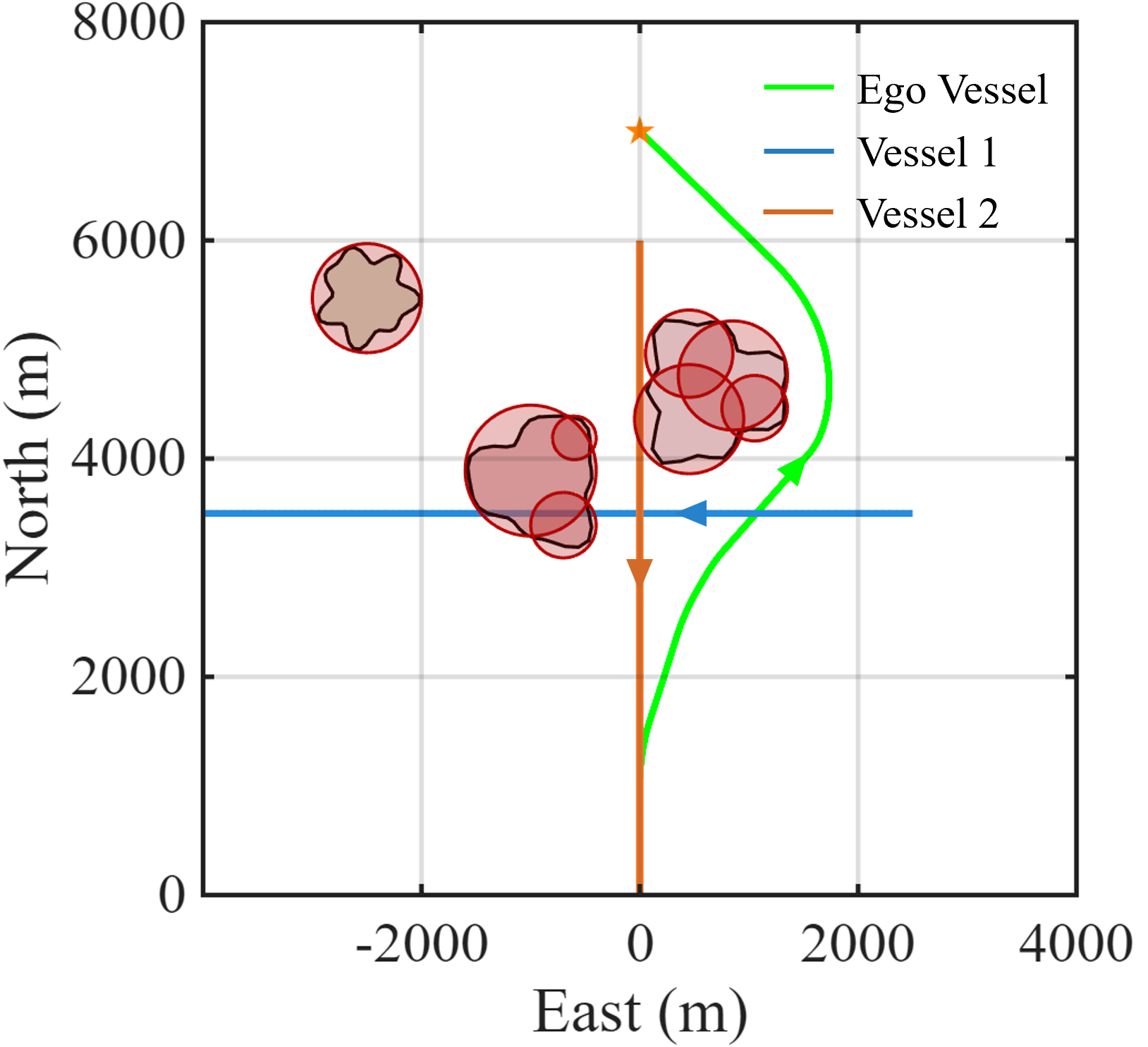}%
        \label{fig:result2}
    }
    \caption{Simulation results for representative encounter scenarios.}
    \label{fig:results_combined}
\end{figure}

\emph{(a) Single head-on encounter:}
Figure~\ref{fig:result1} illustrates the trajectory for the single-vessel head-on case where the ego vessel advances towards the goal while maintaining compliance with  COLREG Rule~14. The bathymetric constraint consisted of a single nonconvex region, approximated using three circles via an ILP formulation, which was solved in 1.75 seconds requiring 81 iterations. The path length traversed was approximately $7840\,\mathrm{m}$, and the vessel retained its cruise speed throughout. The algorithm  maintained a minimum separation of about $552\,\mathrm{m}$ from the target vessel. The computational performance remained efficient, with an average QP solve time of $0.62\,\mathrm{ms}$ and a maximum of $4.83\,\mathrm{ms}$. 


\emph{(b) Head-on and crossing encounter:}
Figure~\ref{fig:result2} shows the configuration involving simultaneous head-on and starboard-side crossing vessels. In this case, the bathymetric constraint consisted of three nonconvex regions, which were approximated by a total of eight circles, and the ILP was solved in 5.48 seconds requiring 151 iterations. The ego vessel traveled approximately $7818\,\mathrm{m}$ during the simulation and maintained safe behavior with a minimum separation of $597\,\mathrm{m}$. Although this scenario triggered a higher number of avoidance constraints, the optimization remained computationally efficient, yielding an average QP solve time of $0.65\,\mathrm{ms}$ and a maximum of approximately $5.79\,\mathrm{ms}$.


Overall, the proposed approach achieved robust and regulation compliant collision avoidance across both encounter scenarios, despite the presence of nonconvex shallow water constraints and dynamic multi-vessel interactions. The ego vessel consistently maintained separation distances that  exceeded $550\,\mathrm{m}$ while complying with the appropriate COLREG maneuvers. Importantly, the underlying QP remained lightweight and computationally efficient, consistently solving within the upper bound of  $6\,\mathrm{ms}$, confirming the real time feasibility of the proposed approach for reliable, efficient, and safe navigation in dynamic maritime environment.


\section{Conclusion}
\label{sec:conclusion}

This paper presented a unified and computationally efficient motion planning method for MASS that simultaneously addresses collision avoidance, COLREGs compliance, and grounding prevention under state uncertainty. The proposed method integrates COLREGs-informed directional constraints, collision constraints based on a robust VO formulation, and convexified bathymetric representations within a single convex programming framework, enabling consistently safe maneuvers under dynamic maritime scenarios.

To incorporate bathymetry constraints, convex and tight over approximations of the highly nonconvex shallow water regions were obtained a priori using a discretized circle-cover formulation solved via ILP. This preprocessing step produces a compact convex map representation suitable for real time motion planning. The simulation results in representative head-on and multi-encounter scenarios demonstrated minimum separation distances that exceeded $550\,\mathrm{m}$ while maintaining consistent regulatory compliance. Across $10{,}000$ control cycles per scenario at a $10\,\mathrm{Hz}$ update rate, the optimization problems were solved within $0.7\,\mathrm{ms}$ on average, with worst case times less than $6\,\mathrm{ms}$, confirming real time feasibility and scalability. In the future, the work will focus on extending the method to large-scale multi-vessel coordination scenarios and incorporating cooperative decision-making strategies. In addition, adaptive refinement of convexified bathymetric representation and integration with higher-level route planning and perception modules will be investigated to enable deployment in complex operational waterways.

\section*{Acknowledgment}
The authors thank American Bureau of Shipping (ABS) for their support. This work was conceptualized as part of a collaborative project with ABS under the Laboratory of Ocean Innovation at Texas A\&M University.

\bibliographystyle{IEEEtran}
\bibliography{references}

@article{khatib1986real,
  title={Real-time obstacle avoidance for manipulators and mobile robots},
  author={Khatib, Oussama},
  journal={The international journal of robotics research},
  volume={5},
  number={1},
  pages={90--98},
  year={1986},
  publisher={Sage Publications Sage CA: Thousand Oaks, CA}
}

@inproceedings{fiorini1993motion,
  title={Motion planning in dynamic environments using the relative velocity paradigm},
  author={Fiorini, Paolo and Shiller, Zvi},
  booktitle={[1993] Proceedings IEEE International Conference on Robotics and Automation},
  pages={560--565},
  year={1993},
  organization={IEEE}
}

@article{fiorini1998motion,
  title={Motion planning in dynamic environments using velocity obstacles},
  author = {Fiorini, P. and Shiller, Zvi},
  journal={The international journal of robotics research},
  volume={17},
  number={7},
  pages={760--772},
  year={1998}
}

@article{johansen2016ship,
  title={Ship collision avoidance using scenario-based model predictive control},
  author={Johansen, Tor A and Cristofaro, Andrea and Perez, Tristan},
  journal={IFAC-PapersOnLine},
  volume={49},
  number={23},
  pages={14--21},
  year={2016},
  publisher={Elsevier}
}

@article{kufoalor2018colregs,
  title={Autonomous collision avoidance for surface vehicles using model predictive control},
  author={Kufoalor, Dela K. M. and Johansen, Tor A. and Brekke, Edmund F.},
  journal={IFAC-PapersOnLine},
  volume={51},
  number={29},
  pages={239--246},
  year={2018},
  publisher={Elsevier}
}

@article{eriksen2019model,
  title={Model predictive control for autonomous ship collision avoidance: A systematic approach to COLREGs compliance},
  author={Eriksen, Bj{\o}rn-Olav H. and Breivik, Morten and Wilthil, Erik F.},
  journal={IEEE Access},
  volume={7},
  pages={157144--157156},
  year={2019},
  publisher={IEEE}
}

@article{campbell2012rule,
  title={A rule-based heuristic method for COLREGS-compliant collision avoidance for an unmanned surface vehicle},
  author={Campbell, Sable and Naeem, Wasif},
  journal={IFAC proceedings volumes},
  volume={45},
  number={27},
  pages={386--391},
  year={2012},
  publisher={Elsevier}
}

@article{penttinen2020colreg,
  title={COLREG compliant collision avoidance using reinforcement learning},
  author={Penttinen, Sebastian},
  year={2020},
  publisher={{\AA}bo Akademi}
}

@article{sawada2021automatic,
  title={Automatic ship collision avoidance using deep reinforcement learning with LSTM in continuous action spaces},
  author={Sawada, Ryohei and Sato, Keiji and Majima, Takahiro},
  journal={Journal of Marine Science and Technology},
  volume={26},
  number={2},
  pages={509--524},
  year={2021},
  publisher={Springer}
}

@article{guan2023generalized,
  title={Generalized behavior decision-making model for ship collision avoidance via reinforcement learning method},
  author={Guan, Wei and Zhao, Ming-yang and Zhang, Cheng-bao and Xi, Zhao-yong},
  journal={Journal of Marine Science and Engineering},
  volume={11},
  number={2},
  pages={273},
  year={2023},
  publisher={MDPI}
}

@inproceedings{kuwata2011safe,
  title={Safe maritime navigation with COLREGS using velocity obstacles},
  author={Kuwata, Yoshiaki and Wolf, Michael T and Zarzhitsky, Dimitri and Huntsberger, Terrance L},
  booktitle={2011 IEEE/RSJ international conference on intelligent robots and systems},
  pages={4728--4734},
  year={2011},
  organization={IEEE}
}

@inproceedings{patil2025virtual,
  title={Virtual Framework for Verification and Validation of Marine Autonomous Navigation},
  author={Patil, Mayur S and Sudharsan, Nataraj and Saaiby, Anthony and Xing, JiaChang and Chan, Jevon and Ammula, Veneela and Tomdio, Jude and Wang, Jin and Rathinam, Sivakumar and Pagilla, Prabhakar and others},
  booktitle={SNAME Maritime Convention},
  pages={D031S019R001},
  year={2025},
  organization={SNAME}
}

@article{eliopoulou2023statistical,
  title={Statistical analysis of accidents and review of safety level of passenger ships},
  author={Eliopoulou, Eleftheria and Alissafaki, Aimilia and Papanikolaou, Apostolos},
  journal={Journal of Marine Science and Engineering},
  volume={11},
  number={2},
  pages={410},
  year={2023},
  publisher={MDPI}
}

@article{gai2024fuzzy,
  title={A Fuzzy Fusion Method for Multi-Ship Collision Avoidance Decision-Making with Merchant and Fishing Vessels},
  author={Gai, Xudong and Zhang, Qiang and Hu, Yancai and Wang, Gang},
  journal={Journal of Marine Science and Engineering},
  volume={12},
  number={10},
  pages={1822},
  year={2024},
  publisher={MDPI}
}

@manual{imo1972colreg,
  title        = {Convention on the International Regulations for Preventing Collisions at Sea, 1972 (COLREG)},
  author       = {{International Maritime Organization}},
  year         = {1972},
  organization = {International Maritime Organization (IMO)},
  address      = {London, UK},
  note         = {Adopted 20 October 1972; entered into force 15 July 1977},
  url          = {https://www.imo.org/en/about/conventions/pages/colreg.aspx}
  }

@unpublished{patil2026robust,
  author = {Patil, Mayur S and Sudharsan, Nataraj and Saaiby, Anthony and others},
  title  = {A Robust Simulation Framework for Verification and Validation of Autonomous Maritime Navigation in Constrained Environments},
  year   = {2026},
  note   = {Manuscript submitted for publication}
}

\end{document}